\documentclass{article}

\usepackage{microtype}
\usepackage{graphicx}
\usepackage{subfigure}
\usepackage{booktabs, multirow}
\usepackage{dsfont}
\usepackage{color}
\usepackage{amsmath,amsfonts,amssymb,bm}
\usepackage{algorithm}
\usepackage{algorithmicx}
\usepackage{algpseudocode}
\usepackage{adjustbox}
\usepackage{balance}

\def\figref#1{Fig.~\ref{#1}}
\def\tabref#1{Table~\ref{#1}}
\def\eqref#1{Eq.~(\ref{#1})}
\def\secref#1{Sec.~\ref{#1}}

\def\sub#1{_{\rm #1}}

\def\eg{{\it e.g.}}

\def\ie{{\it i.e.}}

\newcommand{\argmin}{\mathop{\rm arg~min}\limits}
\algnewcommand\algorithmicforeach{\textbf{for each}}
\algdef{S}[FOR]{ForEach}[1]{\algorithmicforeach\ #1\ \algorithmicdo}
\algblockdefx{FORALLP}{ENDFAP}[1]%
  {\textbf{for all }#1 \textbf{do in parallel}}%
  {\textbf{end for}}

\usepackage{hyperref}

\usepackage[accepted]{icml2021}

\icmltitlerunning{Path Planning using Neural A* Search}

\begin{document}

\twocolumn[
\icmltitle{Path Planning using Neural A* Search}



\icmlsetsymbol{equal}{*}

\begin{icmlauthorlist}
\icmlauthor{Ryo Yonetani}{equal,osx}
\icmlauthor{Tatsunori Taniai}{equal,osx}
\icmlauthor{Mohammadamin Barekatain}{osx,dm}
\icmlauthor{Mai Nishimura}{osx}
\icmlauthor{Asako Kanezaki}{titech}
\end{icmlauthorlist}

\icmlaffiliation{osx}{OMRON SINIC X, Tokyo, Japan}
\icmlaffiliation{titech}{Tokyo Institute of Technology, Tokyo, Japan}
\icmlaffiliation{dm}{Now at DeepMind, London, UK.}

\icmlcorrespondingauthor{Ryo Yonetani}{ryo.yonetani@sinicx.com}

\icmlkeywords{ICML, Path Planning, Neural A* Search}

\vskip 0.3in
]



\printAffiliationsAndNotice{\icmlEqualContribution} 

\begin{abstract}
We present \emph{Neural A*}, a novel data-driven search method for path planning problems. Despite the recent increasing attention to data-driven path planning, machine learning approaches to search-based planning are still challenging due to the discrete nature of search algorithms. In this work, we reformulate a canonical A* search algorithm to be differentiable and couple it with a convolutional encoder to form an end-to-end trainable neural network planner. Neural A* solves a path planning problem by encoding a problem instance to a guidance map and then performing the differentiable A* search with the guidance map. By learning to match the search results with ground-truth paths provided by experts, Neural A* can produce a path consistent with the ground truth accurately and efficiently. Our extensive experiments confirmed that Neural A* outperformed state-of-the-art data-driven planners in terms of the search optimality and efficiency trade-off. Furthermore, Neural A* successfully predicted realistic human trajectories by directly performing search-based planning on natural image inputs\footnote{Project page: \url{https://omron-sinicx.github.io/neural-astar/}.}.
\end{abstract}

\section{Introduction}
\label{sec:intro}

Path planning refers to the problem of finding a valid low-cost path from start to goal points in an environmental map. \emph{Search-based} planning, including the popular A* search \cite{hart1968formal}, is a common approach to path planning problems and has been used in a wide range of applications such as autonomous vehicle navigation~\cite{paden2016survey}, robot arm manipulation~\cite{smith2012dual}, and game AI~\cite{abd2015comprehensive}. Compared to other planning approaches such as sampling-based planning~\cite{gonzalez2015review} and reactive planning \cite{tamar2016value,lee2018gated}, search-based planning is guaranteed to find a solution path, if one exists, by incrementally and extensively exploring the map.

Learning how to plan from expert demonstrations is gathering attention as promising extensions to classical path planners. Recent work has demonstrated major advantages of such data-driven path planning in two scenarios: (1)~finding near-optimal paths more efficiently than classical heuristic planners in point-to-point shortest path search problems~\cite{choudhury2018data,qureshi2019motion,takahashi2019learning,chen2019learning,ichter2020learned} and (2)~enabling path planning on raw image inputs~\cite{tamar2016value,lee2018gated,ichter2019robot,vlastelica2019differentiation}, which is hard for classical planners unless semantic pixel-wise labeling of the environment is given.

\begin{figure}[t]
\centering
    \includegraphics[width=.89\linewidth]{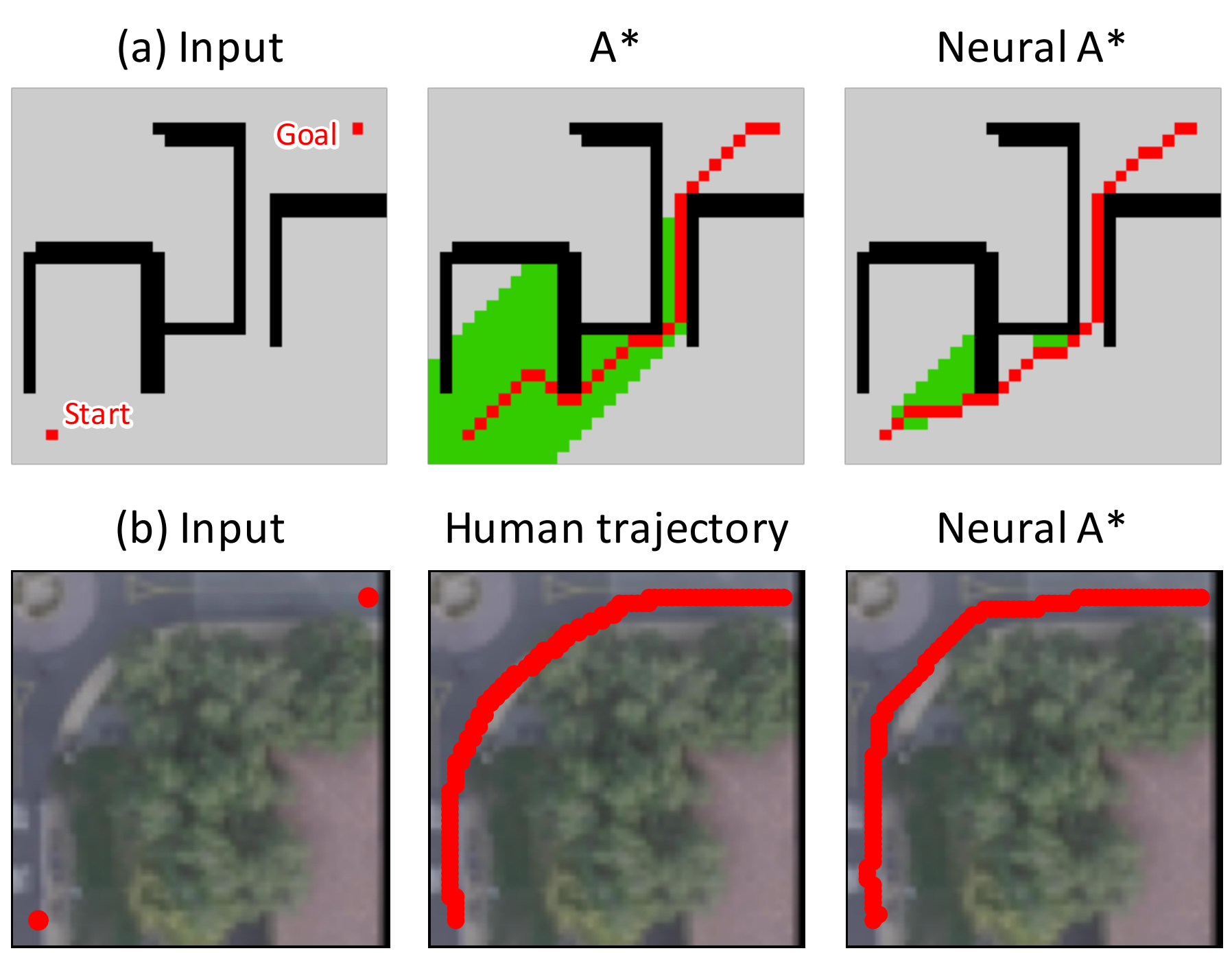}
    \caption{\textbf{Two  Scenarios of Path Planning  with Neural A*.} (a)~Point-to-point shortest path search: finding a near-optimal path (red) with fewer node explorations (green) for an input map. (b)~Path planning on raw image inputs:~accurately predicting a human trajectory (red) on a natural image.}
    \label{fig:teaser}
\end{figure}

In this study, we address both of these separately studied scenarios in a principled fashion, as highlighted in Fig.~\ref{fig:teaser}. In contrast to most of the existing data-driven methods that extend either sampling-based or reactive planning, we pursue a search-based approach to data-driven planning with the intrinsic advantage of guaranteed planning success. 

\begin{figure*}[t]
\centering
    \includegraphics[width=\linewidth]{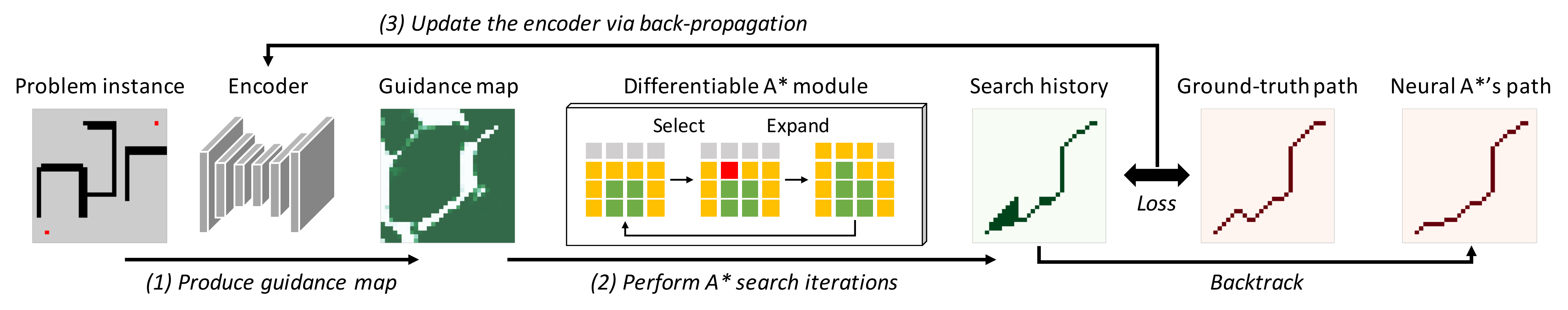}
    \caption{\textbf{Schematic Diagram of Neural A*.} (1) A path-planning problem instance is fed to the encoder to produce a guidance map. (2)~The differentiable A* module performs a point-to-point shortest path search with the guidance map and outputs a search history and a resulting path. (3) A loss between the search history and the ground-truth path is back-propagated to train the encoder.}
    \label{fig:architecture}
\end{figure*}

Studies in this direction so far are largely limited due to the difficulty arising from the discrete nature of incremental search steps in search-based planning, which makes the learning using back-propagation non-trivial. 
Some existing methods thus train heuristic cost functions at each grid point independently, which requires fine-grained expert annotations such as oracle planners running online~\cite{choudhury2018data} and exhaustive pre-computations of the optimal heuristic function of a planner~\cite{takahashi2019learning}. However, such rich annotations are not always available nor scalable, especially when involving labor-intensive manual annotation processes as in~\cite{kim2016socially,kretzschmar2016socially,perez2018learning}. 
More recently, \citet{vlastelica2019differentiation} have applied black-box optimization to combinatorial solvers integrated into neural networks, thus enabling end-to-end learning through combinatorial algorithms including search-based planning. This general-purpose optimization can be adapted to our problem. However, treating the entire search procedure as a black-box function loses the detailed track of internal search steps, making the training hard.

To address the non-differentiability of search-based planning, we propose a novel data-driven search-based planner named Neural A*. At its core, \emph{we reformulate a canonical A* search algorithm to be differentiable} as a module referred to as the differentiable A*, by combining a discretized activation technique inspired by \citet{hubara2016binarized} with basic matrix operations. This module enables us to perform 
an A* search in the forward pass of a neural network and back-propagate losses through every search step to other trainable backbone modules. As illustrated in \figref{fig:architecture}, Neural A* consists of the combination of a fully-convolutional encoder and the differentiable A* module, and is trained as follows: \mbox{(1)} Given a problem instance  (\ie, an environmental map annotated with start and goal points), the encoder transforms it into  a scalar-valued map representation referred to as a guidance map; \mbox{(2)} The differentiable A* module then performs a search with the guidance map to output a search history and a resulting path; \mbox{(3)} The search history is compared against the ground-truth path of the input instance to derive a loss, which is back-propagated to train the encoder.

The role of the differentiable A* for training is simple: teaching the encoder to produce guidance maps that lead to minimizing the discrepancy between the resulting search histories and ground-truth paths.
The encoder then learns to capture visual cues in the inputs that are effective for reproducing the ground-truth paths. This learning principle provides unified solutions to the aforementioned two problem scenarios. Specifically, in the shortest-path search scenario (Fig.~\ref{fig:teaser}a), where the ground-truth paths are given by optimal planners, the encoder is trained to find near-optimal paths efficiently by exploiting visual cues such as shapes of dead ends. Here, guidance maps are served to augment the input maps to prioritize which nodes to explore/avoid for improving the search optimality-efficiency trade-off.
On the other hand, when the ground-truth paths are given by human annotators for raw image inputs (Fig.~\ref{fig:teaser}b), the encoder enables the planning directly on the images by learning passable and impassable regions from colors and textures as low- and high-cost nodes in guidance maps.

We extensively evaluate our approach on both synthetic~\cite{bhardwaj2017learning} and real-world datasets~\cite{sturtevant2012benchmarks,robicquet2016learning}. Our results show that Neural A* outperformed state-of-the-art data-driven search-based planners~\cite{choudhury2018data,vlastelica2019differentiation} in terms of the trade-off between search optimality and efficiency for point-to-point shortest path problems. Further, we demonstrate that Neural A* can learn to predict pedestrian trajectories from raw real-world surveillance images more accurately than \citet{vlastelica2019differentiation} and other imitation learners~\cite{ratliff2006max,tamar2016value,lee2018gated}.

\section{Preliminaries}
\paragraph{Path planning problem.}
Let us consider a path planning problem, in particular a point-to-point shortest (\ie, lowest-cost) path problem, on a graph $\mathcal{G}=(\mathcal{V}, \mathcal{E})$ where $\mathcal{V}$ is the set of nodes representing the locations of the environment and $\mathcal{E}$ is the set of potentially valid movements between nodes. For convenience, we define an alternate graph form $\mathcal{G} = (\mathcal{V}, \mathcal{N})$ with $\mathcal{N}(v)=\{v'\mid (v, v')\in \mathcal{E}, v\neq v'\}$ referring to the set of the neighbor nodes of $v$. Each edge $(v, v')$ is given a non-negative movement cost $c(v')\in\mathbb{R}_+$ that depends only on the node $v'$. Given start and goal nodes, $v\sub{s}, v\sub{g}\in\mathcal{V}$, the objective of path planning is to find a sequence of connected nodes, $P=(v_1, v_2,\dots,v_T)\in\mathcal{V}^T$, $v_1=v\sub{s}$, $v_T=v\sub{g}$, with its total cost $\sum_{t=1}^{T-1}c(v_{t+1})$ being the lowest among all possible paths from $v\sub{s}$ to $v\sub{g}$. Following \citet{choudhury2018data}, this work focuses on a popular setting where $\mathcal{G}$ refers to the eight-neighbor grid world and $c(v')$ is a unit cost taking $c(v')=1$ when $v'$ is passable and $c(v')=\infty$ otherwise, \eg, when $v'$ is occupied by an obstacle.

\paragraph{A* search.}
\begin{figure}[t]
  \begin{algorithm}[H]
    \caption{A* Search}
    \label{alg:va}
    \begin{algorithmic}[1]
      \Require Graph $\mathcal{G}$, movement cost $c$, start $v\sub{s}$, and goal $v\sub{g}$
      \Ensure Shortest path $P$
      \State Initialize $\mathcal{O}\leftarrow v\sub{s}$, $\mathcal{\mathcal{C}}\leftarrow\emptyset$, $\texttt{Parent}(v\sub{s})\leftarrow \emptyset$.
      \While {$v\sub{g} \notin \mathcal{C}$}
          \State Select $v^*\in\mathcal{O}$ based on \eqref{eq:select_node}.
          \State Update $O\leftarrow O \setminus v^*$, $\mathcal{C}\leftarrow \mathcal{C}\cup v^*$.
          \State Extract $\mathcal{V}\sub{nbr}\subset\mathcal{V}$ based on \eqref{eq:expand_nodes}.
          \ForEach {$v'\in\mathcal{V}\sub{nbr}$}
          \State Update $\mathcal{O}\leftarrow \mathcal{O} \cup v'$, $\texttt{Parent}(v') \leftarrow v^*$.
          \EndFor
    \EndWhile
          \State $P\leftarrow \texttt{Backtrack}(\texttt{Parent}, v\sub{g})$.
    \end{algorithmic}
  \end{algorithm}
\end{figure}

Algorithm~\ref{alg:va} overviews the implementation of A* search employed in this work. It explores nodes to find a shortest path $P$ by iterating between (1) selecting the most promising node from the list of candidates for constructing a shortest path and (2) expanding neighbors of the selected node to update the candidate list, until the goal $v\sub{g}$ is selected. More specifically, the node selection (Line 3 of Algorithm~\ref{alg:va}) is done based on the following criterion:
\begin{equation}
    v^* = \argmin_{v\in\mathcal{O}}\left(g(v) + h(v)\right),
    \label{eq:select_node}
\end{equation}
where $\mathcal{O}\subset\mathcal{V}$ is an \emph{open list} that manages candidate nodes for the node selection. $g(v)$ refers to the actual total cost accumulating $c(v')$ for the nodes $v'$ along the current best path from $v\sub{s}$ to $v$, which is updated incrementally during the search. On the other hand, $h(v)$ is a heuristic function estimating the total cost from $v$ to $v\sub{g}$, for which the straight-line distance between $v$ and $v\sub{g}$ is often used in the grid world. All the selected nodes are stored in another list of search histories called \emph{closed list}, $\mathcal{C}\subseteq\mathcal{V}$, as done in Line~4.

In Line 5 of Algorithm~\ref{alg:va}, we expand the neighboring nodes of $v^*$ as $\mathcal{V}\sub{nbr}\subset\mathcal{V}$ based on the following criterion:
\begin{equation}
\mathcal{V}\sub{nbr}=\{v'\mid v'\in \mathcal{N}(v^*) \land v' \notin \mathcal{O} \land v'\notin \mathcal{\mathcal{C}}\}.
\label{eq:expand_nodes}
\end{equation}
The neighbor nodes $\mathcal{V}\sub{nbr}$ are then added to $\mathcal{O}$ in Line 7 to propose new selection candidates in the next iteration. The search is terminated once $v\sub{g}$ is selected in Line 3 and stored in $\mathcal{C}$, followed by $\texttt{Backtrack}$ that traces the parent nodes $\texttt{Parent}(v)$ from $v\sub{g}$ to $v\sub{s}$ to obtain a solution path $P$.

\paragraph{Data-driven planning setup.}
As explained in \secref{sec:intro}, we seek a principled search-based approach to two distinct scenarios of data-driven path planning in existing work: (1) finding near-optimal paths efficiently for point-to-point shortest path problems~\cite{choudhury2018data}, and (2) enabling the path planning on raw images in the absence of movement costs~\cite{vlastelica2019differentiation}.

To this end, we abstract these two settings by introducing a 2D environmental-map variable $X$ representing the input graph $\mathcal{G}$. 
Specifically, corresponding to the two settings with known and unknown movement costs $c(v')$, the map $X$ represents $c(v')$ either (1) explicitly as a binary map $X\in\{0, 1\}^\mathcal{V}$ taking $1$ for passable locations (\ie, $c(v')=1$) and $0$ otherwise, or (2) implicitly as a raw colored image $X\in[0, 1]^{3 \times \mathcal{V}}$. As a result, each problem instance is represented consistently as an indexed tuple $Q^{(i)}=(X^{(i)}, v\sub{s}^{(i)}, v\sub{g}^{(i)})$.

For each problem instance $Q^{(i)}$, we further assume that a ground-truth path is given as a 2D binary map $\bar{P}^{(i)}\in\{0, 1\}^\mathcal{V}$ whose elements take $1$ along the desired path. When $X^{(i)}$ is a binary map indicating the movement cost explicitly, $\bar{P}^{(i)}$ is obtained by solving the shortest path problem using an optimal planner. If $X^{(i)}$ is a raw image, we assume that $\bar{P}^{(i)}$ is provided by a human annotator.
\section{Neural A* Search}
Now we present the proposed Neural A* search. At a high level, Neural A* uses an encoder to transform a problem instance $Q^{(i)}$ into a \emph{guidance map}, as illustrated in \figref{fig:architecture}. This guidance map imposes a \emph{guidance cost} $\phi^{(i)}(v)\in\mathbb{R}_+$ to each node $v$. Then, the differentiable A* module, detailed in \secref{subsec:differentiable_astar}, performs a search under the policy to minimize a total guidance cost, \ie, $\sum_{t=1}^{T-1}\phi^{(i)}(v_{t+1})$. By repeating this forward pass and the back-propagation through the training procedure described in \secref{subsec:training}, the encoder learns to capture visual cues in the training instances to improve the path accuracy and search efficiency.

\subsection{Differentiable A* Module} 
\label{subsec:differentiable_astar}
\paragraph{Variable representations.} To reformulate A* search in a differentiable fashion, we represent the variables in Algorithm~\ref{alg:va} as matrices of the map size so that each line can be executed via matrix operations. Specifically, let $O, C, V\sub{nbr} \in \mathcal\{0, 1\}^\mathcal{V}$ be binary matrices indicating the nodes contained in $\mathcal{O}, \mathcal{C}, \mathcal{V}\sub{nbr}$, respectively (we omit an index $i$ here for simplicity.) We represent the start node $v\sub{s}$, goal node $v\sub{g}$, and selected node $v^*$ as one-hot indicator matrices $V\sub{s}, V\sub{g}, V^*\in \{0, 1\}^\mathcal{V}$, respectively, where $\langle V\sub{s}, \mathds{1}\rangle=\langle V\sub{g}, \mathds{1}\rangle = \langle V^*, \mathds{1}\rangle = 1$ (\ie, one-hot), $\langle A, B\rangle$ is the inner product of two matrices $A$ and $B$, and $\mathds{1}$ is the all-ones matrix of the map size. In addition, let $G, H, \Phi \in \mathbb{R}_+^\mathcal{V}$ be a matrix version of $g(v)$, $h(v)$, and $\phi(v)$, respectively.

\paragraph{Node selection.} Performing \eqref{eq:select_node} in a differentiable manner is non-trivial as it involves a discrete operation. Here, we leverage a discretized activation inspired by \citet{hubara2016binarized} and reformulate the equation as follows:
\begin{equation}
V^* = \mathcal{I}_\text{max}\left(\frac{\exp(-(G+H)/\tau) \odot O}{\langle
\exp(-(G+H)/\tau), O \rangle }\right),
\label{eq:select_diff}
\end{equation}
where $A\odot B$ denotes the element-wise product, and $\tau$ is a temperature parameter that will be defined empirically. $\mathcal{I}_\text{max}(A)$ is the function that gives the argmax index of $A$ as a binary one-hot matrix during a forward pass, while acting as the identity function for back-propagation. The open-list matrix~$O$ is used here to mask the exponential term so that the selection is done from the nodes in the current open list.

\paragraph{Node expansion.} Expanding the neighboring nodes of $v^*$ in \eqref{eq:expand_nodes} involves multiple conditions, \ie, $v'\in\mathcal{N}(v^*)$, $v'\notin \mathcal{O}$, and $v'\notin \mathcal{C}$. Inspired by \citet{tamar2016value}, we implement $\mathcal{N}$ as a convolution between $V^*$ and the fixed kernel $K=[[1, 1, 1]^{\top}, [1, 0, 1]^{\top}, [1, 1, 1]^{\top}]$. When $X$ is given as a binary cost map indicating the passable/impassable nodes, $V\sub{nbr}$ is obtained by the following matrix operations:
\begin{equation}
V\sub{nbr} = (V^* * K) \odot X \odot (\mathds{1} - O) \odot (\mathds{1} - C),
\label{eq:expand_diff}
\end{equation}
where $A * B$ is the 2D convolution of $A$ and $B$. The multiplication by $X \odot (\mathds{1} - O) \odot (\mathds{1} - C)$ acts as a mask to extract the nodes that are passable and not contained in the open nor close lists. 
Masking with $X$ preserves the original obstacle structures and thus the graph topology, keeping the differentiable A* \emph{complete} (\ie, guaranteed to always find a solution if one exists in the graph $\mathcal{G}$, as in standard A*). We also introduce $\bar{V}\sub{nbr}=(V^* * K)\odot X \odot O\odot (\mathds{1}-C)$ indicating the neighboring nodes already in the open list, which we will use to update $G$ below. When $X$ is otherwise a raw image that does not explicitly indicate the passable nodes, we use $V\sub{nbr} = (V^* * K) \odot (\mathds{1} - O) \odot (\mathds{1} - C)$ and $\bar{V}\sub{nbr}=(V^* * K)\odot O\odot (\mathds{1}-C)$.

\paragraph{Updating $G$.} As explained earlier, $g(v)$ here represents the total guidance cost paid for the actual path from $v\sub{s}$ to $v$, instead of accumulating the original movement cost $c(v')$ that is not always available explicitly. To update this total cost at each iteration, we partially update $G$ with new costs $G'$ for the neighbor nodes as follows:
\begin{eqnarray}
G &\leftarrow& G' \odot V\sub{nbr} +  \min(G', G)\odot \bar{V}\sub{nbr} \nonumber\\
&&+ G \odot (\mathds{1} - V\sub{nbr} - \bar{V}\sub{nbr}), 
\label{eq:g}\\
G' &=&\langle G, V^*\rangle\cdot \mathds{1} + \Phi.
\label{eq:g2}
\end{eqnarray}
In \eqref{eq:g}, the neighbors $V\sub{nbr}$ that are opened for the first time are assigned $G'$, and the neighbors $\bar{V}\sub{nbr}$ that have already been opened are assigned the lower of the new costs $G'$ and previously computed costs $G$.
The new costs $G'$ for the neighbors are computed in \eqref{eq:g2} as the sum of the current cost of the selected node, $g(v^*)$, expressed using $g(v^*)=\langle G, V^*\rangle$, and the one-step guidance cost to the neighbors represented by $\Phi$.

\subsection{Training Neural A*}
\label{subsec:training}
\paragraph{Loss design.} 
The differentiable A* module connects its guidance-map input $\Phi$ and search output so that a loss for the output is back-propagated to $\Phi$ and, consequently, to the encoder. Here, the output is the closed list $C$, which is a binary matrix accumulating all the searched nodes $V^*$ from \eqref{eq:select_diff} in a search (see \eqref{eq:oc_mlt} for details). We evaluate the mean $L_1$ loss between $C$ and the ground-truth path map $\bar{P}$:
\begin{equation}
    \mathcal{L} = \| C - \bar{P} \|_1 / |\mathcal{V}|.
    \label{eq:loss}
\end{equation}
This loss supervises the node selections by penalizing both (1)~the false-negative selections of nodes that should have been taken into $C$ to find $\bar{P}$ and (2)~the false-positive selections of nodes that were excessive in $C$ to match with $\bar{P}$. In other words, this loss encourages Neural A* to (1)~search for a path that is close to the ground-truth path (2)~with fewer node explorations.

In practice, we disable the gradients of $\mathcal{O}$ in \eqref{eq:select_diff}, $V\sub{nbr}, \bar{V}\sub{nbr}$ in \eqref{eq:g}, and $G$ in \eqref{eq:g2} by detaching them from back-propagation chains. Doing so effectively informs the encoder of the importance of the  guidance cost $\Phi$ in \eqref{eq:g2} for the node selection in \eqref{eq:select_diff}, while simplifying recurrent back-propagation chains to stabilize the training process and reduce large memory consumption\footnote{We found in our experiments that fully enabling the gradients caused training failures due to an out-of-memory problem with 16GB GPU RAM.}.

\paragraph{Encoder design.} 
The loss shown above is back-propagated through every search step in the differentiable A* module to the encoder. Here, we expect the encoder to learn visual cues in the given problem instances for enabling accurate and efficient search-based planning. These cues include, for instance, the shapes of dead ends and bypasses in binary cost maps or colors and textural patterns of passable roads in raw natural images. For this purpose, we utilize a fully-convolutional network architecture such as U-Net~\cite{ronneberger2015u} used for semantic segmentation, which can learn local visual representations at the original resolution useful for the downstream task\footnote{As the nature of convolutional neural networks, guidance cost outputs are only sensitive to the input map within their limited receptive fields. Thus, when the map size is intractably large, we could partially sweep the input map with the encoder incrementally during a search without changing search results.}. 
The input to the encoder is given as the concatenation of $X$ and $V\sub{s} + V\sub{g}$. In this way, the extraction of those visual cues is properly conditioned by the start and goal positions.

\paragraph{Enabling mini-batch training.}
The complete algorithm of Neural A* is summarized in Algorithm~\ref{alg:na}. To accelerate the training, it is critical to process multiple problem instances at once in a mini batch. However, this is not straightforward because those intra-batch samples may be solved within different numbers of search steps. We address this issue by introducing a binary goal verification flag $\eta^{(i)} = 1 - \langle V^{(i)}\sub{g}, V^{*(i)}\rangle$ and updating $O^{(i)},C^{(i)}$ as follows:
\begin{equation}
O^{(i)} \leftarrow O^{(i)} - \eta^{(i)} V^{*(i)},\;\; C^{(i)} \leftarrow C^{(i)} + \eta^{(i)} V^{*(i)}.
\label{eq:oc_mlt}
\end{equation}
Intuitively, these operations keep $O^{(i)}$ and $C^{(i)}$ unchanged once the goal is found. We repeat Lines 9--15 until we obtain $\eta^{(i)} = 0$ for all the samples in the batch.

\begin{figure}[t]
  \begin{algorithm}[H]
    \caption{Neural A* Search}
    \label{alg:na}
    \begin{algorithmic}[1]
      \Require Problem instances $\{Q^{(i)}=(X^{(i)}, v^{(i)}\sub{s}, v^{(i)}\sub{g})\mid i=1,\dots,b\}$ in a mini-batch of size $b$.
      \Ensure Closed-list matrices $\{C^{(i)}\mid i =1,\dots,b\}$ and solution paths $\{P^{(i)}\mid i=1,\dots,b\}$.
      \FORALLP {$i=1,\dots,b$}
      \State Compute $V^{(i)}\sub{s},V^{(i)}\sub{g}$ from $v^{(i)}\sub{s}, v^{(i)}\sub{g}$.
      \State Compute $\Phi^{(i)}$ from $X^{(i)}, V^{(i)}\sub{s}, V^{(i)}\sub{g}$ by the encoder.
      \State Initialize $O^{(i)}\leftarrow V^{(i)}\sub{s}$, $C^{(i)}\leftarrow \mathbf{0}$, $G^{(i)}\leftarrow \mathbf{0}$.
      \State Initialize $\texttt{Parent}^{(i)}(v^{(i)}\sub{s})\leftarrow \emptyset$.
      \ENDFAP
      \Repeat
      \FORALLP {$i=1,\dots,b$}
          \State Select $V^{*(i)}$ based on \eqref{eq:select_diff}.
          \State Compute $\eta^{(i)} = 1 - \langle V^{(i)}\sub{g}, V^{*(i)}\rangle$.
          \State Update $O^{(i)}$ and $C^{(i)}$ based on \eqref{eq:oc_mlt}.
          \State Compute
          $V^{(i)}\sub{nbr}$ based on \eqref{eq:expand_diff}.
          \State Update $O^{(i)}\leftarrow O^{(i)} + V^{(i)}\sub{nbr}$.
          \State Update $G^{(i)}$ based on \eqref{eq:g} and \eqref{eq:g2}.
          \State Update $\texttt{Parent}^{(i)}$ based on Algorithm~\ref{alg:va}-L6,7.
    \ENDFAP
    \Until $\eta^{(i)}=0$ for $i=1,\dots,b$
      \FORALLP {$i=1,\dots,b$}
    \State $P^{(i)}\leftarrow \texttt{Backtrack}(\texttt{Parent}^{(i)}, v^{(i)}\sub{g})$.
    \ENDFAP
    \end{algorithmic}
  \end{algorithm}
\end{figure}
\section{Experiments}
In this section, we first conduct an extensive experiment to evaluate the effect of Neural A* on the search optimality and efficiency trade-off, \ie, how efficiently Neural A* search can find near-optimal paths for point-to-point shortest path problems. Due to space limitations, we provide the detailed experimental setups in Appendix~\ref{app:dataset}.

\subsection{Datasets}
We adopted the following public path-planning datasets with obstacle annotations to collect planning demonstrations.
\begin{itemize}
\item \textbf{Motion Planning (MP) Dataset}: A collection of eight types of grid-world environments with distinctive obstacle shapes, created by \citet{bhardwaj2017learning}. Each environment group consists of 800 training, 100 validation, and 100 test maps, with the same type of obstacles placed in different layouts. We resized each environmental map into the size of $32\times 32$ to complete the whole experiment in a reasonable time. By following the original setting, the training and evaluation were conducted for each environment group independently.
\item \textbf{Tiled MP Dataset}: Our extension to the MP dataset to make obstacle structures more complex and diverse. Specifically, we tiled four maps drawn randomly from the MP dataset to construct an environmental map with the size of $64\times64$. We repeated this process to create 3,200 training, 400 validation, and 400 test maps.
\item \textbf{City/Street Map (CSM) Dataset}: A collection of 30 city maps with explicit obstacle annotations as binary images, organized by \citet{sturtevant2012benchmarks}. For data augmentation, we cropped multiple random patches with the size of $128\times 128$ from each map and resized them to the size of $64\times 64$. We used 20 of the 30 maps to generate random 3,200 training and 400 validation maps and the remaining 10 maps for 400 test samples. In this way, we ensured that no maps were shared between training/validation and test splits. 
\end{itemize}

For each map, we created problem instances by randomly picking out a single goal from one of the four corner regions of the map as well as one, six, and fifteen start locations for training, validation, and test splits, respectively, from areas sufficiently distant from the goal. We obtained the ground-truth shortest paths using the Dijkstra algorithm.

\subsection{Methods and Implementations}

\paragraph{Neural A*.} As the encoder, we adopted U-Net~\cite{ronneberger2015u} with the VGG-16 backbone~\cite{simonyan2014very} implemented by \citet{Yakubovskiy:2019}, where the final layer was activated by the sigmoid function to constrain guidance maps to be $\Phi\in[0, 1]^\mathcal{V}$. For the differentiable A* module, we used the Chebyshev distance as the heuristic function $H$ suitable for the eight-neighbor grid-world~\cite{patel1997heuristics}. The Euclidean distance multiplied by a small constant ($0.001$) was further added to $H$ for tie-breaking. The temperature $\tau$ in \eqref{eq:select_diff} was empirically set to the square root of the map width. 

\paragraph{Baselines.}
For assessing the significance of Neural A* in search-based planning, we adopted the following data-driven search-based planners as baseline competitors that have the planning success guarantee, like ours, by design. We extended the authors' implementations available online.
\begin{itemize}
\item \textbf{SAIL}~\cite{choudhury2018data}: A data-driven best-first search method that achieved high search efficiency by learning a heuristic function from demonstrations. Unlike Neural A*, SAIL employs hand-engineered features such as the straight-line distances from each node to the goal and the closest obstacles. We evaluated two variants of SAIL where training samples were rolled out using the learned heuristic function (\textbf{SAIL}) or the oracle planner (\textbf{SAIL-SL}).
\item \textbf{Blackbox Differentiation}~\cite{vlastelica2019differentiation}: A general-purpose approach to data-driven combinatorial solvers including search-based path planning. It consists of an encoder module that transforms environmental maps into node-cost maps and a solver module that performs a combinatorial algorithm as a piece-wise constant black-box function taking the cost maps as input. We implemented a Blackbox-Differentiation version of A* search (\textbf{BB-A*}) by treating our differentiable A* module as a black-box function and training the encoder via black-box optimization, while keeping other setups, such as the encoder architecture and loss function, the same as those of Neural A*.
\end{itemize}
We also evaluated best-first search (\textbf{BF}) and weighted A* search (\textbf{WA*})~\cite{pohl1970heuristic} as baseline classical planners, which used the same heuristic function as that of Neural A*. Finally, we implemented a degraded version of Neural A* named \textbf{Neural BF}, which always used $G=\Phi$ in \eqref{eq:select_diff} instead of $G$ incrementally updated by \eqref{eq:g}. By doing so, Neural BF ignored the accumulation of guidance costs from the start node, much like best-first search.

\subsection{Experimental Setups}
All the models were trained using the RMSProp optimizer, with the mini-batch size, the number of epochs, and the learning rate set to (100, 100, 0.001) for the MP dataset and (100, 400, 0.001) for the Tiled MP and CSM datasets.

For each trained model, we calculated the following metrics to evaluate how much the trade-off between search optimality and efficiency was improved from a standard A* search performed using the identical heuristic function.
\begin{itemize}
    \item \textbf{Path optimality ratio (Opt)} measuring the percentage of shortest path predictions for each environmental map, as used by \citet{vlastelica2019differentiation}.
    \item \textbf{Reduction ratio of node explorations (Exp)} measuring the number of search steps reduced by a model compared to the standard A* search in $0-100$ (\%). Specifically, by letting $E^*$ and $E$ be the number of node explorations by A* and a model, respectively, it is defined by $\max(100 \times \frac{E^*-E}{E^*}, 0)$ and averaged over all the problem instances for each environmental map.
    \item \textbf{The Harmonic mean (Hmean)} of Opt and Exp, showing how much their trade-off was improved by a model.
\end{itemize}
During the training, we saved model weights that marked the best Hmean score on the validation split and used them for measuring final performances on the test split. To investigate statistical differences among models, we computed the bootstrap mean and 95\% confidence bounds per metric.

\subsection{Results}
\begin{table}[t]
\caption{\textbf{Quantitative Results.} Bootstrap means and 95\% confidence bounds of path optimality ratio (Opt), reduction ratio of node explorations (Exp), and their harmonic mean (Hmean).}
\label{tab:result_all}
\centering
\scalebox{0.8}{
\begin{tabular}{lccc}
\toprule 
\multicolumn{4}{c}{\textsc{MP Dataset}} \\
\midrule
 & Opt & Exp & Hmean\\ 
  \midrule
 BF & 65.8 (63.8, 68.0) &  44.1 (42.8, 45.5) &  44.8 (43.4, 46.3) \\
 WA * & 68.4 (66.5, 70.4) & 35.8 (34.5, 37.1) & 40.4 (39.0, 41.8) \\
\midrule
SAIL& 34.6 (32.1, 37.0) &  \textbf{48.6 (47.2, 50.2)} &  26.3 (24.6, 28.0) \\
SAIL-SL & 37.2 (34.8, 39.5) &  46.3 (44.8, 47.8) &  28.3 (26.6, 29.9) \\
BB-A* & 62.7 (60.6, 64.9) &  42.0 (40.6, 43.4) &  42.1 (40.5, 43.6) \\
\midrule
Neural BF &  75.5 (73.8, 77.1) &  45.9 (44.6, 47.2) &  \textbf{52.0 (50.7, 53.4)} \\
\textbf{Neural A*} & \textbf{87.7 (86.6, 88.9)} &  40.1 (38.9, 41.3) &  \textbf{52.0 (50.7, 53.3)} \\
\midrule
\midrule
\multicolumn{4}{c}{\textsc{Tiled MP Dataset}} \\
\midrule
  & Opt & Exp & Hmean\\ 
  \midrule
BF & 32.3 (30.0, 34.6) &  58.9 (57.1, 60.8) &  34.0 (32.1, 36.0) \\
WA* &  35.3 (32.9, 37.7) & 52.6 (50.8, 54.5) & 34.3 (32.5, 36.1) \\
\midrule
SAIL& 5.3 (4.3, 6.1) & 58.4 (56.6, 60.3)  & 7.5 (6.3, 8.6)  \\
SAIL-SL & 6.6 (5.6, 7.6)&  54.6 (52.7, 56.5) & 9.1 (7.9, 10.3) \\
BB-A* & 31.2 (28.8, 33.5) &  52.0 (50.2, 53.9) &  31.1 (29.2, 33.0) \\
\midrule
Neural BF & 43.7 (41.4, 46.1) &  \textbf{61.5 (59.7, 63.3)} &  44.4 (42.5, 46.2) \\
\textbf{Neural A*} &\textbf{63.0 (60.7, 65.2)} &  55.8 (54.1, 57.5) &  \textbf{54.2 (52.6, 55.8)} \\
\midrule
\midrule
\multicolumn{4}{c}{\textsc{CSM Dataset}} \\
\midrule
   & Opt & Exp & Hmean\\ 
  \midrule
BF & 54.4 (51.8, 57.0) &  39.9 (37.6, 42.2) &  35.7 (33.9, 37.6) \\
WA* & 55.7 (53.1, 58.3) &  37.1 (34.8, 39.3) & 34.4 (32.6, 36.3) \\
\midrule
SAIL& 20.6 (18.6, 22.6) & 41.0 (38.8, 43.3) & 18.3 (16.7, 19.9) \\
SAIL-SL & 21.4 (19.4, 23.3) & 39.3 (37.1, 41.6) &  17.6 (16.1, 19.1) \\
BB-A* & 54.4 (51.8, 57.1) &  40.0 (37.7, 42.3) &  35.6 (33.8, 37.4) \\
\midrule
Neural BF & 60.9 (58.5, 63.3) &  \textbf{42.1 (39.8, 44.3)} &  40.6 (38.7, 42.6) \\
\textbf{Neural A*} & \textbf{73.5 (71.5, 75.5)} &  37.6 (35.5, 39.7) &  \textbf{43.6 (41.7, 45.5)} \\
\bottomrule
\end{tabular}
}
\end{table}

\begin{figure*}[t]
    \centering
    \includegraphics[width=\linewidth]{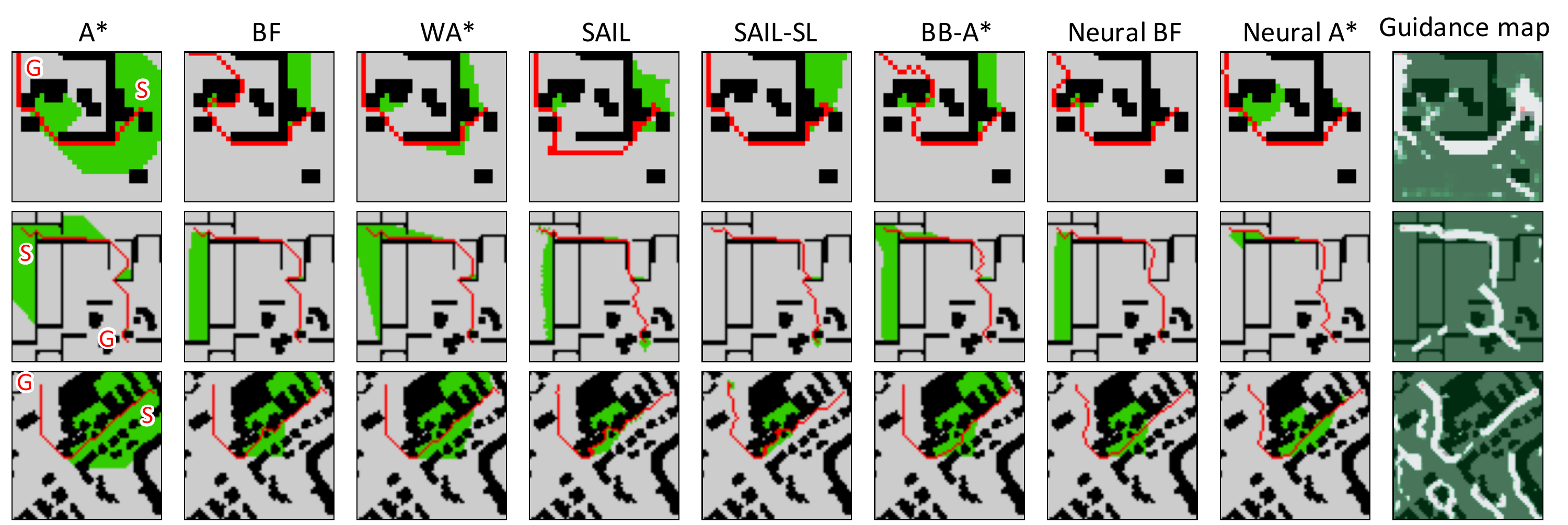}
	\caption{\textbf{Selected Path Planning Results.} Black pixels indicate obstacles. Start nodes (indicated by ``S''), goal nodes (indicated by ``G''), and found paths are annotated in red. Other explored nodes are colored in green. In the rightmost column, guidance maps are overlaid on the input maps where regions with lower costs are visualized in white. More results are presented in Appendix~\ref{app:additional_results}.}
    \label{fig:results}
\end{figure*}

\paragraph{Comparisons with baselines.}
Table~\ref{tab:result_all} summarizes quantitative results. Overall, Neural A* outperformed baseline methods in terms of both Opt and Hmean and improved the path optimality and search efficiency trade-off. SAIL and SAIL-SL sometimes performed more efficiently, which however came with low optimality ratios especially for larger and more complex maps of the Tiled MP and CSM datasets. BB-A* showed higher Hmean scores than those of SAIL/SAIL-SL but was consistently outperformed by Neural A*. Since the only difference between Neural A* and BB-A* is whether making the A* search differentiable or treating it as a black-box in updating an identically-configured encoder, the comparison between them highlights the effectiveness of our approach. With the differentiable A* module, Neural A* has access to richer information of internal steps, which is necessary to effectively analyze causal relationships between individual node selections by \eqref{eq:select_diff} and results. This information is however black-boxed in BB-A*. 
Also, we found that classical heuristic planners (BF and WA*) performed comparably to or sometimes better than other data-driven baselines. This result could be explained by our challenging experimental setup adopting randomized start and goal locations instead of pre-defined ones by the original work~\cite{choudhury2018data,vlastelica2019differentiation}. Finally, Neural BF achieved the highest Exp scores in the Tiled MP and CSM datasets but often ended up with sub-optimal paths since it ignored the actual cost accumulation. For more evaluations including analysis on the complexity and runtime of Neural A*, see Appendix~\ref{app:additional_results}.

\begin{figure}[t]
    \centering
    \includegraphics[width=\linewidth]{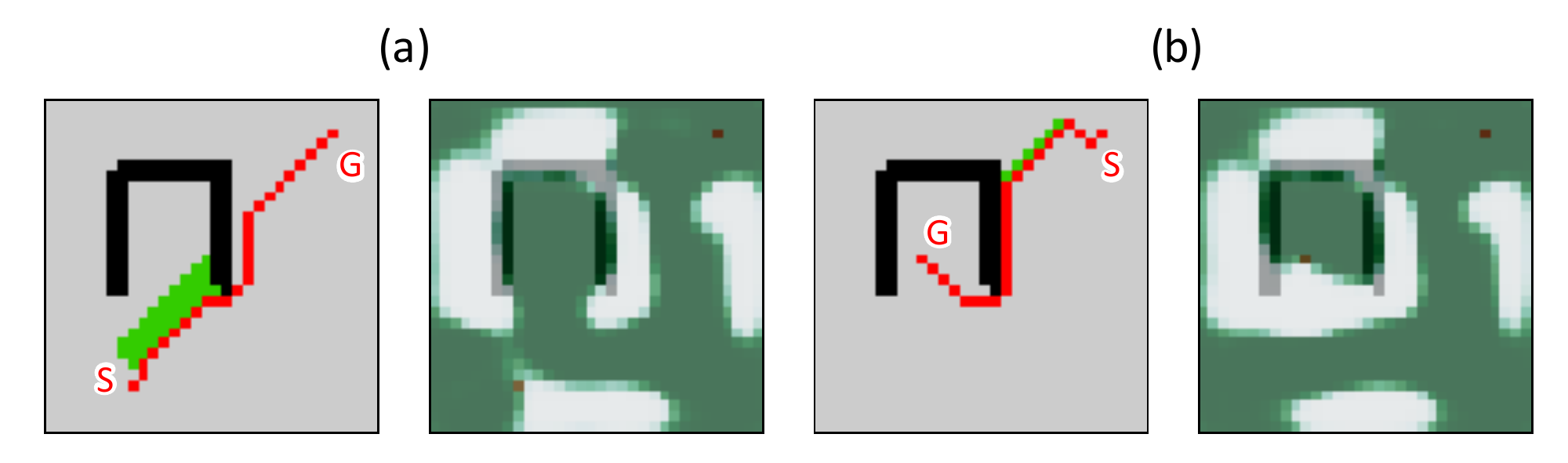}
    \caption{\textbf{Adaptive Encoding Results.} (a) The U-shaped dead-end obstacle is placed between the start and goal nodes; (b) The goal node is located inside the U-shaped dead end.}
    \label{fig:ablation}
\end{figure}

\paragraph{Qualitative results.} Figure~\ref{fig:results} visualizes example search results with guidance maps produced by the encoder of Neural A*. We confirmed that the encoder successfully captured visual cues in problem instances. Specifically, higher guidance costs (shown in green) were assigned to the whole obstacle regions creating dead ends, while lower costs (shown in white) were given to bypasses and shortcuts that guided the search to the goal. Figure~\ref{fig:ablation} depicts how Neural A* performed a search adaptively for different start and goal locations in the same map. Notice the entrance to the U-shaped dead end is (a) blocked with 
high guidance costs when the dead end is placed between the start and goal and (b) open when the goal is inside the dead end.

\paragraph{Ablation study.} We further assessed the effects of the encoder by comparing several different configurations for its architecture. In the configuration \textbf{w/o start and goal}, we fed only an environmental map $X_i$ to the encoder as input. 
In \textbf{w/ ResNet-18 backbone}, a residual network architecture~\cite{he2016deep} was used instead of VGG-16.
Table \ref{tab:ablation} shows the degraded performances with these configurations especially in terms of the optimality ratio.
Additionally, we tried the following variants of the proposed approach, which we found not effective. Adopting the straight-through Gumbel-softmax~\cite{jang2016categorical} for \eqref{eq:select_diff} caused complete planning failures (with Opt, Exp, Hmean all $0$) as it biases the planner to avoid the most promising nodes.
Using the mean squared loss for \eqref{eq:loss} produces the same gradients as the mean $L_1$ loss up to a constant factor for binary variables (\ie, $C$ and $\bar{P}$), and indeed led to a comparable performance.

\begin{table}[t]
\caption{\textbf{Ablation Study.} Performance comparisons with different architectural design choices for the encoder on the MP dataset.}
\label{tab:ablation}
\centering
\scalebox{0.65}{
\begin{tabular}{lccc}
\toprule 
   & Opt & Exp & Hmean\\ 
  \midrule
\textbf{Neural A*} & \textbf{87.7 (86.6, 88.9)} &  40.1 (38.9, 41.3) &  \textbf{52.0 (50.7, 53.3)} \\
\midrule
w/o start and goal &  67.0 (65.1, 68.8) & 36.8 (35.6, 38.1) & 41.5 (40.2, 42.7) \\
w/ ResNet-18 backbone &  79.8 (78.1, 81.5) & \textbf{41.4 (40.2, 42.7)} & 49.2 (47.9, 50.5)\\
\bottomrule
\end{tabular}
}
\end{table}

\paragraph{Limitations.}
Although Neural A* worked well in various environments, its current implementation assumes the grid world environments with unit node costs. One interesting direction is to extend Neural A* to work on a high-dimensional or continuous state space. This will require the encoder to be tailored to such a space, as done in~\citet{qureshi2019motion,chen2019learning,ichter2019robot}. We leave this extension for future work.

\section{Path Planning on Raw-Image Inputs}
As another scenario for Neural A*, we address the task of planning paths directly on raw image inputs. Specifically, suppose a video of an outdoor scene taken by a stationary surveillance camera. Planning demonstrations then consist of color images of the scene and actual trajectories of pedestrians (\ie, ground-truth paths provided by human annotators). Given these data, we aim by Neural A* to predict realistic trajectories consistent with those of pedestrians when start and goal locations are provided.
We here compared Neural A* with BB-A*~\cite{vlastelica2019differentiation} as a competitor that can also perform planning on raw image inputs. For more comparisons with other imitation learners~\cite{ratliff2006max,tamar2016value,lee2018gated}, see Appendix~\ref{app:il_methods}.

\paragraph{Dataset.}
We used Stanford Drone Dataset (SDD)~\cite{robicquet2016learning}, which comprises surveillance videos captured by static drone cameras placed at eight distinct locations. We split the dataset in two ways: (1) intra-scenes: for each location, choosing one video to build a single test split while using the rest to train a model, to see if the model can predict trajectories of unseen individuals observed at different times; (2) inter-scenes: performing leave-one-location-out cross-validation to see how well a learned model can generalize to unseen locations. As planning demonstrations, we extracted pedestrian trajectories and the local patch of a video frame that encompassed the trajectories.

\paragraph{Implementations and experimental setups.} Unlike the previous experiment with obstacle locations given explicitly, each model now has to learn visual representations of obstacles to avoid them during node selections. Therefore, we modified the U-Net encoder by multiplying the final sigmoid activation by a trainable positive scalar (initialized to $10.0$) so that obstacle regions can be assigned sufficiently high costs. We trained both Neural A* and BB-A* using RMSProp with the batch size, the number of epochs, and the learning rate set to $(64, 20, 0.001)$. Because the main objective of this task is to predict paths close to the ground-truth ones, rather than to improve the search optimality and efficiency trade-off, we calculated the chamfer distance as a metric for evaluating the dissimilarities between those paths.

\begin{figure}[t]
    \centering
    \includegraphics[width=\linewidth]{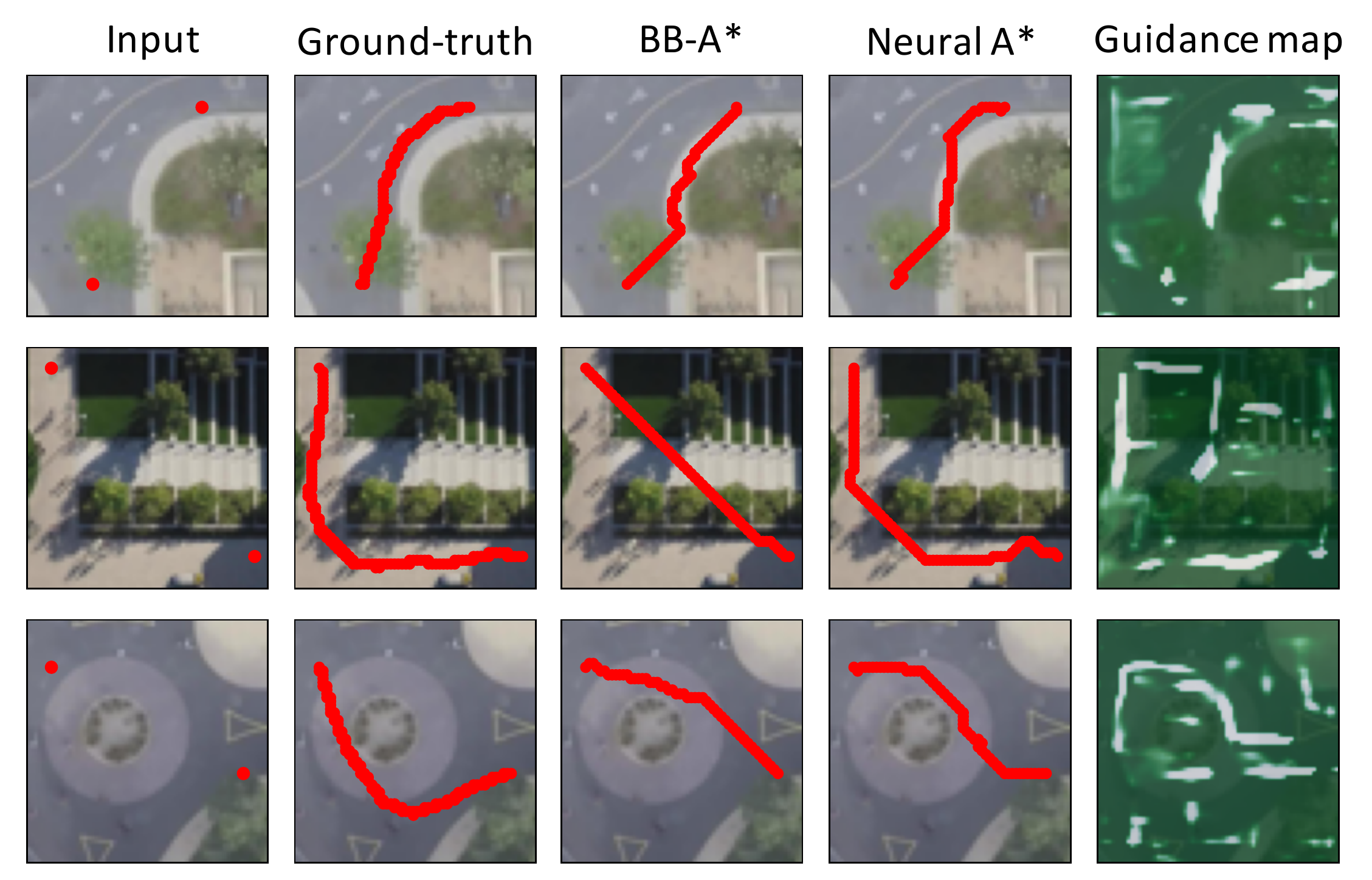}
    \caption{\textbf{Path Planning Examples on SDD.} Neural A* predicted paths similar to actual pedestrian trajectories.}
    \label{fig:result_sdd}
\end{figure}

\paragraph{Results.} 
Table~\ref{tab:sdd} shows that Neural A* significantly outperformed \mbox{BB-A*}. As visualized in the first two examples in \figref{fig:result_sdd}, Neural A* often predicted paths along roads, resulting in predictions closer to ground-truth paths compared to those by BB-A*. However, both methods sometimes failed to predict actual pedestrian trajectories when there were multiple possible routes to the destinations, as shown in the example at the bottom. A possible extension to address this issue is to adopt a generative framework~\cite{gupta2018social,salzmann2020trajectron++} that allows multiple paths to be stochastically sampled.

\begin{table}[t]
\caption{\textbf{Quantitative Results on SDD.} Bootstrap means and 95\% confidence bounds of the chamfer distance between predicted and ground-truth pedestrian trajectories.}
\label{tab:sdd}
\centering
\scalebox{0.9}{
\begin{tabular}{lcc}
    \toprule 
  & Intra-scenes & Inter-scenes\\ 
  \midrule
BB-A* & 152.2 (144.9, 159.1) & 134.3 (132.1, 136.4)\\
\textbf{Neural A*} & \textbf{16.1 (13.2, 18.8)} & \textbf{37.4 (35.8, 39.0)} \\
\bottomrule
\end{tabular}
}
\end{table}

\section{Related Work}
Approaches to path planning problems can be broadly classified into several types, each with its advantages and limitations. For example, sampling-based planning, such as RRT~\cite{lavalle1998rapidly} and PRM~\cite{kavraki1996probabilistic}, can explore high-dimensional state spaces and has widely been used for robot motion planning~\cite{kingston2018sampling}. 
A practical challenge is identifying important regions in the state spaces to effectively sample sparse state-points for efficient planning.
To this end, a variety of data-driven methods have been developed to learn such regions~\cite{ichter2018learning,ichter2019robot} or to learn exploration strategies~\cite{qureshi2019motion,perez2018learning,chen2019learning} from expert demonstrations or prior successful planning experiences.

Meanwhile, reactive planners learn a policy of moving agents to determine the next best actions, such as moving left or right, from current states via supervised learning~\cite{tamar2016value,kanezaki2017goselo,gupta2017cognitive,karkus2017qmdp,lee2018gated,bency2019neural} or inverse reinforcement learning (IRL)~\cite{ratliff2006max,ziebart2008maximum,wulfmeier2015maximum,kim2016socially,kretzschmar2016socially,fu2017learning,yu2019meta}. These approaches can be useful for dynamic environments where the agents have to act adaptively. Also, IRL-based methods are relevant to our work in that they learn to recover cost functions from demonstrations. However, they often suffer from planning failures especially for a large map and thus require the help of other classical planners to enable long-range planning~\cite{faust2018prm}. In Appendix~\ref{app:il_methods}, we provide quantitative evaluations of some relevant methods~\cite{ratliff2006max,tamar2016value,lee2018gated} with our experimental setting.

Compared to these two approaches, search-based planning is advantageous in terms of ensured success in finding valid paths in a fine grid map. Classical heuristic planners have sought better heuristic functions, search algorithms, and efficient implementations to improve the search performance (\eg, \citet{burns2012implementing,zhou2015massively,abd2015comprehensive}). Recent data-driven approaches further extend heuristic planners in two ways: learning from expert demonstrations to (a) find near-optimal paths efficiently~\cite{choudhury2018data,takahashi2019learning} or (b) perform the planning directly on raw image inputs~\cite{vlastelica2019differentiation}. Standing on both sides, our work proposes the first differentiable search-based planner that can solve these two problems in a principled fashion.
From another perspective, data-driven heuristic planners can learn their cost functions, as in ours and \citet{vlastelica2019differentiation}, as well as their heuristic functions, as in \citet{choudhury2018data,takahashi2019learning}. These variations are analogous to learning a (negative) reward function in IRL and a Q-function in RL, respectively. Very recently, \citet{archetti2021neural} have extended Neural A* to learn both of these functions.

\section{Conclusion}
We have presented a novel data-driven search-based planner named Neural A*, which involves a differentiable A* algorithm. Neural A* learns from demonstrations to improve the trade-off between search optimality and efficiency in path planning and also to enable the planning directly on raw image inputs. Our extensive experimental evaluations on multiple public datasets have demonstrated the effectiveness of our approach over state-of-the-art planners.


\bibliographystyle{icml2021}
\balance

\appendix
\section{Details of Experimental Setups}
\label{app:dataset}

\subsection{Dataset Creation}
Here we present supplementary information on how each dataset was created in our experiments. Since our dataset generation process involves randomness, we fixed a random seed in each generation program to ensure reproducibility. Please refer to the code in our project page: \url{https://omron-sinicx.github.io/neural-astar/}.

\begin{figure}[t]
    \centering
    \includegraphics[width=\linewidth]{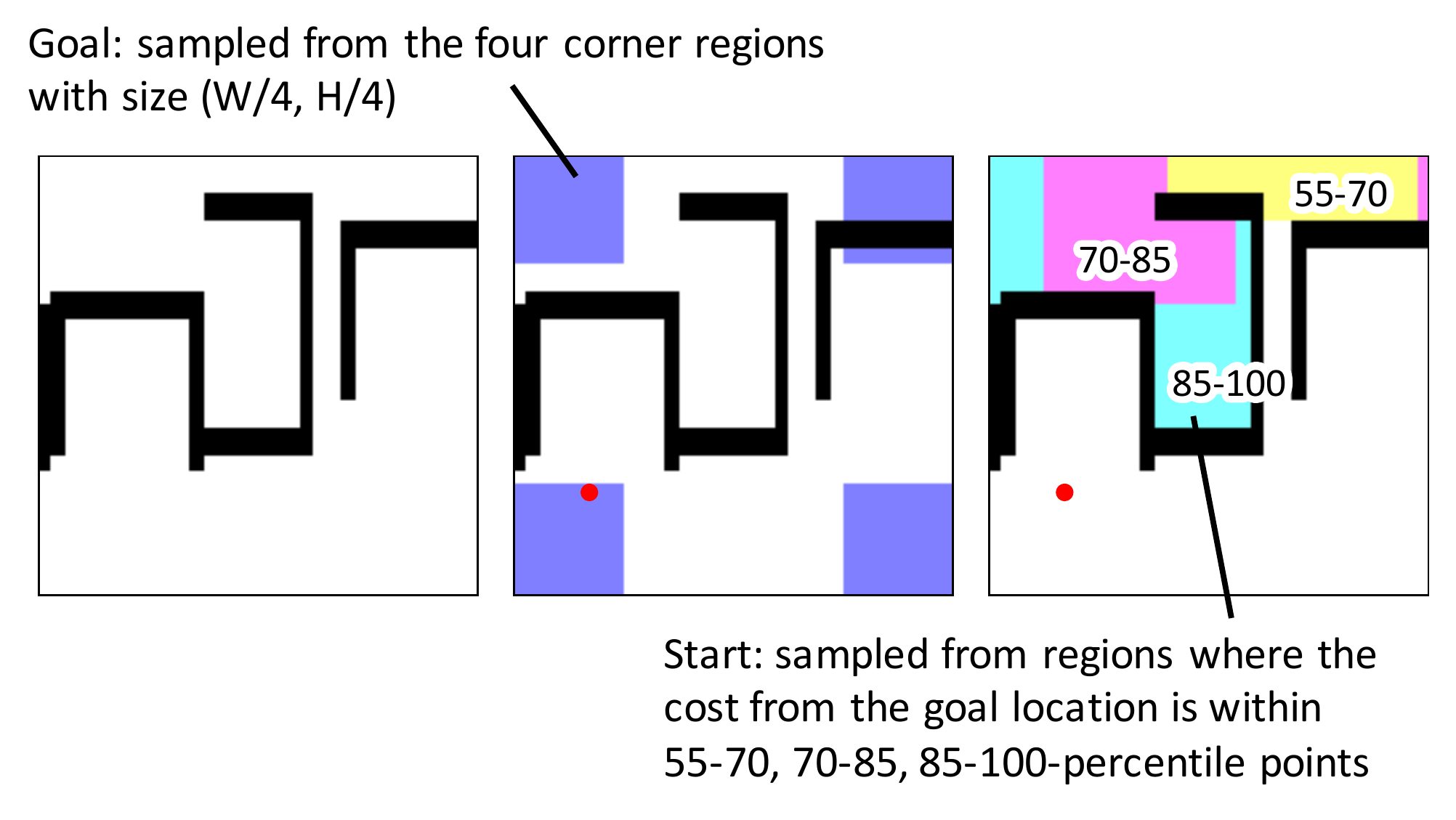}
    \vskip -2mm
    \caption{\textbf{Sampling of Start and Goal Locations}.}
    \label{fig:start_goal}
\end{figure}

\paragraph{MP/Tiled-MP/CSM datasets.}   
In the experiments with MP/Tiled-MP/CSM datasets, we employed more challenging settings involving randomized start and goal locations instead of pre-defined consistent locations used in prior work~\cite{choudhury2018data,vlastelica2019differentiation}. 
We determined these start and goal locations strategically based on their actual distances to avoid generating easy problem instances.
Specifically, for each environmental map, a single goal location was randomly determined once and fixed throughout the experiments. Here, for a map with the width and height denoted as $(W, H)$, \ie, $(32, 32)$ for the MP and $(64, 64)$ for the Tiled MP and CSM datasets, the goal location was sampled from one of the four corner regions of size $(W/4, H/4)$, as illustrated in \figref{fig:start_goal} (middle). Then, we performed the Dijkstra algorithm to compute actual movement costs from every passable node to the goal, and calculated the $55, 70, 85$-percentile points of the costs. For every iteration in the training phase, we sampled a new random start location from regions whose actual costs higher than the $55$ percentile point. As for validation and testing data, we sampled two and five random but fixed start locations, respectively, from each of three kinds of regions whose costs are within the percentile ranges of $[55, 70]$, $[70, 85]$, and $[85, 100]$ (see \figref{fig:start_goal} (right) for an illustration). Consequently, we created $2\times3=6$ and $5\times3 = 15$ diverse start locations for each validation and test map, respectively. The ground-truth paths for all the generated problem instances were obtained by the Dijkstra algorithm. When computing losses for the Tiled MP and CSM datasets, these paths were dilated with a $3\times 3$ kernel, which stabilized the training. 

\paragraph{SDD.}
In SDD, we extracted relatively simple trajectories of pedestrians who moved directly towards their destinations. Specifically, for each trajectory provided by \citet{robicquet2016learning}, we first trimmed it randomly to have a sequence length in the range of $[300, 600]$ timesteps (at 2.5fps). We then calculated the ratio of its straight-line distance between start and goal points to the trajectory length, as a measure of path simplicity that gives a lower value for a more complex trajectory. We discarded trajectories whose simplicity was less than 0.5. Finally, we cropped image patches that encompass each trajectory with the margin of 50 pixels and resized them to the size of $64\times 64$. As a result, 8,325 samples were extracted from the dataset.

\begin{table}[t]
\caption{\textbf{Hyper-parameter Selection.} The list of hyper-parameters and ranges of these values tried during the development of this paper.}
\label{tab:parameter}
\centering
\scalebox{0.75}{
\begin{tabular}{lc}
    \toprule 
  Parameter Name & Values (range of values tried)\\ 
  \midrule
    \multicolumn{2}{c}{\textbf{Common}}\\ 
    \midrule
  Optimizer & RMSProp \\
  Learning rate & 0.001 (0.0001, 0.0005, 0.001, 0.005)\\
  Batch size & 100 \\
  Number of epochs & 100 (MP), 400 (Tiled MP, CSM) \\
    Tie breaking (for $h$ func.) & $0.001\times$ Euclidean distance \\
  \midrule
    \midrule
    \multicolumn{2}{c}{\textbf{Neural A*/Neural BF}}\\ 
    \midrule
    \shortstack{\vspace{0.5em}Encoder arch} & \shortstack{U-Net with VGG-16 backbone\\(VGG-16, ResNet-18)} \\
    Temperature $\tau$ & $\sqrt{32}$ for MP, $\sqrt{64}$ for Tiled MP and CSM\\
  \midrule
    \midrule
    \multicolumn{2}{c}{\textbf{BB-A*}}\\ 
    \midrule
    Encoder arch & U-Net with VGG-16 backbone \\
    Trade-off parameter $\lambda$ & 20.0 (0.1, 1.0, 5.0, 10.0, 20.0) \\ 
  \midrule
    \midrule
    \multicolumn{2}{c}{\textbf{SAIL/SAIL-SL}}\\ 
    \midrule
    Max data samples & 300 (60, 300) \\
    Sampling rate & 10 (10, 100) \\
  \midrule
    \midrule
    \multicolumn{2}{c}{\textbf{WA*}}\\ 
    \midrule
    Weight factor for $h(v)$ & 0.8 (0.6, 0.7, 0.8) \\
\bottomrule
\end{tabular}
}
\end{table}

\subsection{Hyper-Parameter Selection}
Table~\ref{tab:parameter} shows the list of hyper-parameters as well as ranges of these values we tried to produce the final results. We selected the final parameters based on the validation performance on the MP dataset in terms of Hmean scores. Because completing all the experiments took a considerably long time (see the next section), we performed each experiment only once with a fixed set of random seeds.

Below we provide several remarks regarding the hyper-parameter list.
We observed that the tie-breaking in A* search, \ie, the adjustment to $h(v)$ by adding the Euclidean distance from $v$ to the goal scaled by a small positive constant (0.001), was critical to improving the base efficiency of A* search. Thus, we used this setting for all the A*-based methods throughout the experiments. The choice of the learning rate little affected final performances given a sufficient number of epochs. BB-A* has an additional hyper-parameter $\lambda$ that controls the trade-off between ``informativeness of the gradient'' and ``faithfulness to the original function''~\cite{vlastelica2019differentiation}. We tried several values and found that any choice worked reasonably well, except for extremely small values (\eg, 0.1). SAIL and SAIL-SL~\cite{choudhury2018data} have hyper-parameters on how to collect samples from each training environment instance, which little affected final performances. Finally, the weighted A* baseline used a modified node selection criterion with a weight factor $w$ to the heuristic function; \ie, $(1-w)\cdot g(v) + w \cdot h(v)$, for which we set $w=0.8$ throughout the experiments. Note that using $w=1.0$ for the criterion corresponds to the best-first search in our baselines.

\subsection{Computing Infrastructure and Training Time}
\label{app:infra}
We performed all the experiments on a server operated on Ubuntu 18.04.3 LTS with NVIDIA V100 GPUs, Intel Xeon Gold 6252 CPU @ 2.10GHz (48 cores), and 768GB main memory. Our implementation was based on PyTorch 1.5~\cite{NEURIPS2019_9015} and Segmentation Models PyTorch~\cite{Yakubovskiy:2019}. To efficiently carry out the experiments, we used GNU Parallel~\cite{Tange2011a} to run multiple experiment sessions in parallel. See our \texttt{setup.py} and \texttt{Dockerfile} for the list of all dependencies.

In the training sessions, each model was trained using a single V100 GPU with 16 GB graphics memory. Training each of our models (Neural A* and Neural BF) took approximately 50 minutes on the MP dataset (100 epochs $\times$ 800 maps with the size of $32\times32$) and 35 hours on the Tiled MP and CSM datasets (400 epochs $\times$ 3,200 maps with the size of $64\times64$). As for SAIL/SAIL-SL and BB-A*, the training on the MP dataset took approximately the same time (\ie, 50 minutes). On the Tiled MP and CSM datasets, SAIL/SAIL-SL took up to about 22 hours and BB-A* took 65 hours.

\section{Additional Results}
\label{app:additional_results}

\begin{figure*}[t]
    \centering
    \includegraphics[width=.84\linewidth]{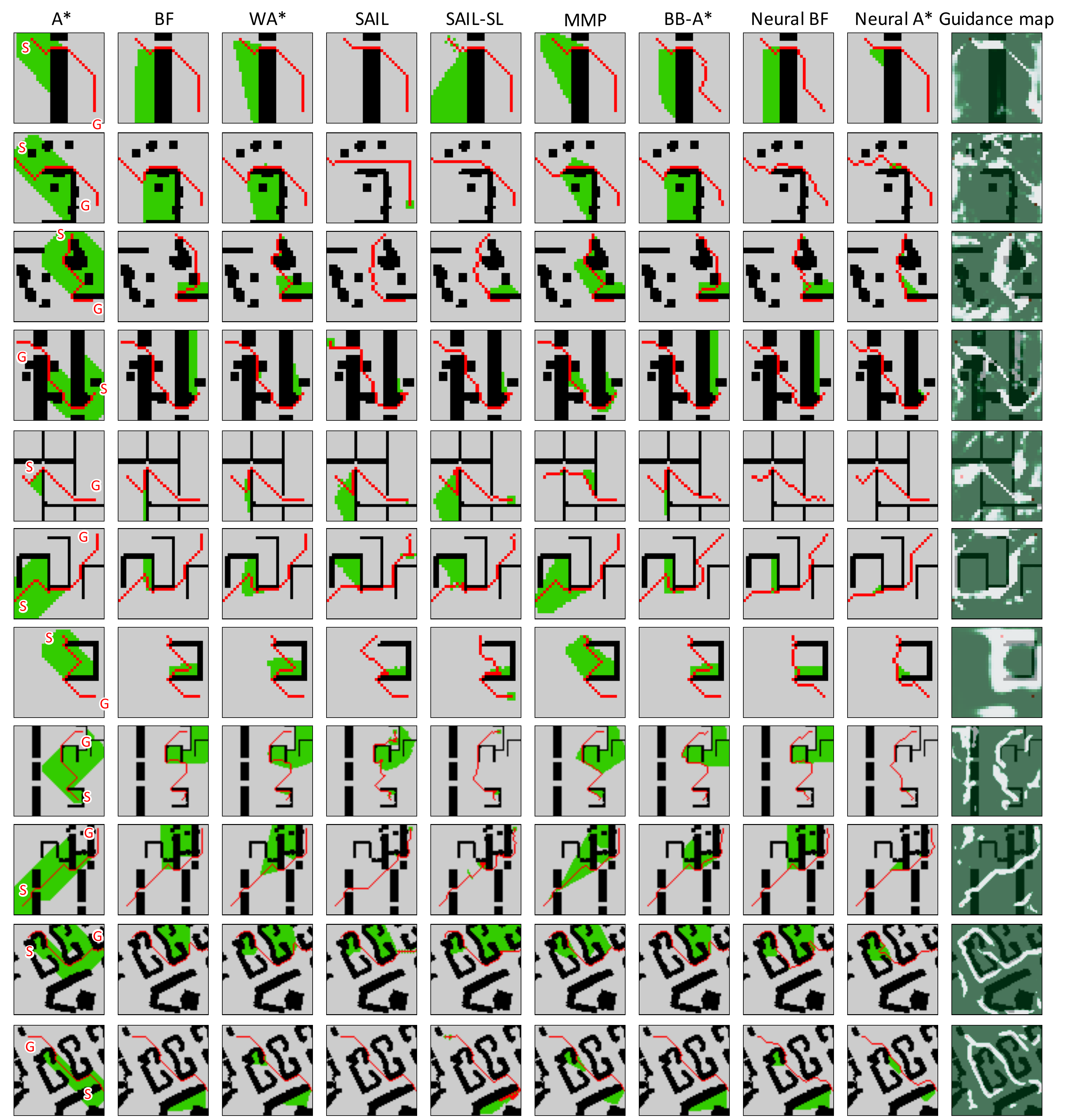}
    \caption{\textbf{Additional Qualitative Results (MP/Tiled-MP/CSM Datasets)}.}
    \label{fig:results_extra}
\end{figure*}

\begin{figure*}[t]
    \centering
    \includegraphics[width=0.7\linewidth]{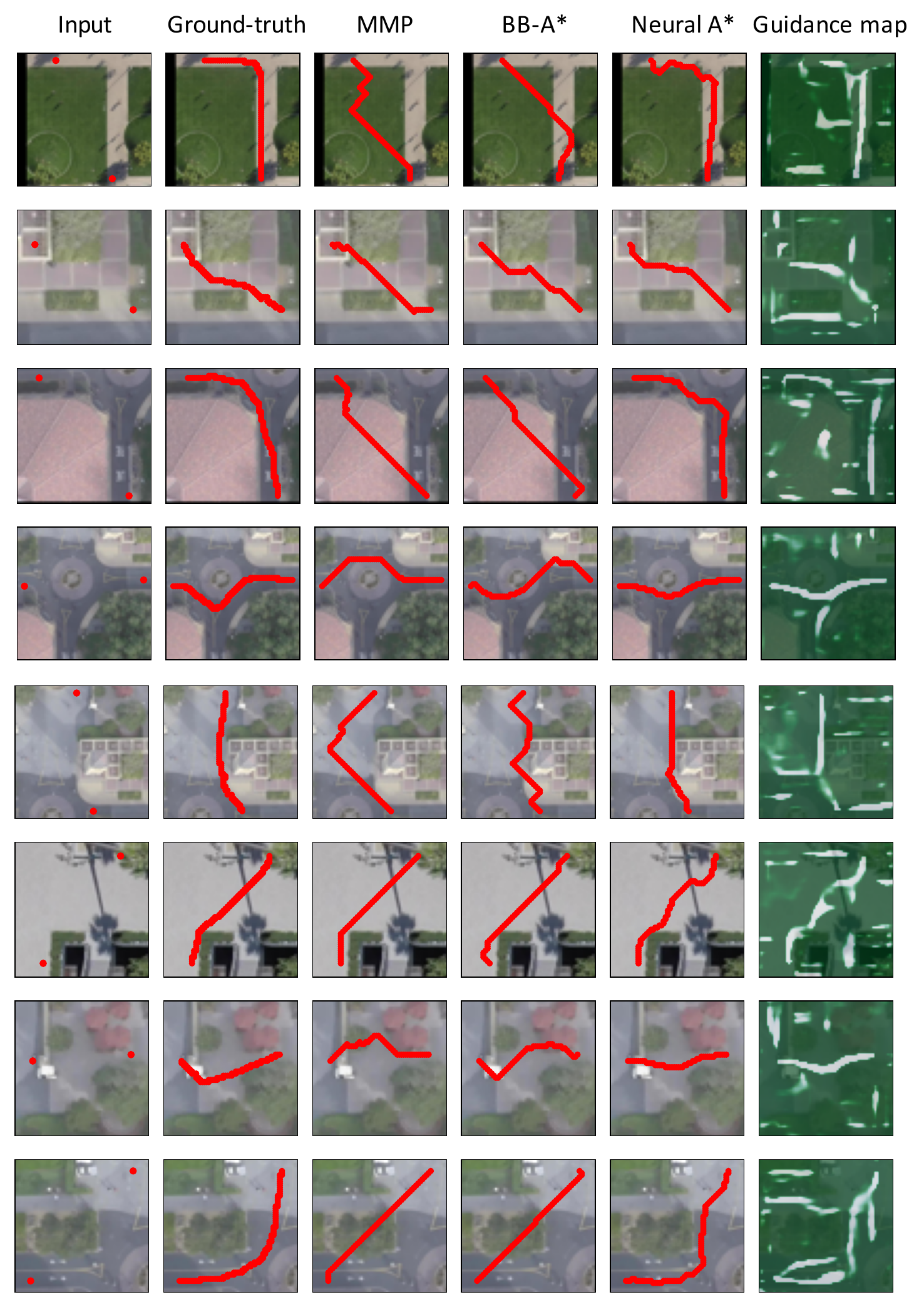}
    \caption{\textbf{Additional Qualitative Results (SDD)}.}
    \label{fig:results_sdd_extra}
\end{figure*}

Figures~\ref{fig:results_extra} and \ref{fig:results_sdd_extra} show additional qualitative results (including those of MMP introduced below.) In what follows, we also add more detailed performance analysis to Neural A* from different perspectives.

\subsection{Comparisons with Imitation Learning Methods}
\label{app:il_methods}
As introduced in Sec.~6, some of the imitation learning (IL) methods are relevant to our work in that they learn to recover unknown reward (\ie, negative cost) functions from demonstrations.
Here, we compared our approach with several IL methods tailored to path planning tasks, namely, Maximum Margin Planning (MMP)~\cite{ratliff2006max}, Value Iteration Network (VIN)~\cite{tamar2016value} and Gated Path Planning Network (GPPN)~\cite{lee2018gated}. 

For MMP, we followed the original work and modeled the cost function as a per-node linear mapping from features to costs. For feature extraction, we used the VGG-16 network pretrained on ImageNet. As in Neural A*, we activated the costs with the sigmoid function to constrain them to be in $[0, 1]$. Unlike other IL-based planners, MMP uses A* search to plan a path with the estimated cost. We used the same implementation of A* search as that of Neural A*.

For VIN and GPPN, we used the official codebase of \citet{lee2018gated}. Because these reactive planners are not based on a search-based algorithm, we could not employ the Exp and Hmean metrics, which are associated with performances of the baseline A* search.
Instead, we calculated the success ratio (Suc), which is the percentage of problem instances for which a planner found a valid path.

As shown in Tables~\ref{tab:result_extra} and \ref{tab:sdd_extra}, we confirmed that the performances of these IL methods are limited compared to the proposed Neural A*. Although MMP ensures 100\% planning success as using A* search, it was consistently outperformed by Neural A* in terms of the Opt, Exp, and Hmean metrics. One possible reason of these limited performances is that MMP cannot learn how internal search steps of A* affect search histories and resulting paths, as we compared BB-A* against Neural A* in Sec.~4.4. While GPPN obtained a higher Opt score than Neural A* on the Tiled MP dataset, it did not always successfully find a valid path as shown in its success ratio. Moreover, GPPN and VIN completely failed to learn on SDD. These failures can be accounted for by its large input maps and limited demonstrations (single trajectory per map), which are more challenging settings than those by the original work.

\begin{table*}[p]
\parbox{0.75\textwidth}{\caption{\textbf{Comparisons with Imitation Learning Methods.} Bootstrap means and 95\% confidence bounds of path optimality ratio (Opt), reduction ratio of node explorations (Exp), the harmonic mean (Hmean) between Opt and Exp, and success ratio (Suc).}
\label{tab:result_extra}}
\centering
\scalebox{0.9}{
\begin{tabular}{lcccc}
\toprule 
\multicolumn{5}{c}{\textsc{MP Dataset}} \\
\midrule
 & Opt & Exp & Hmean & Suc\\ 
\midrule
VIN & 24.7 (23.9, 25.5) & N/A & N/A & 31.4 (30.6, 32.3) \\
GPPN & 71.0 (70.2, 71.8) & N/A & N/A & 86.2 (85.6, 86.9)\\
MMP & 81.6 (80.6, 82.8) & 22.4 (21.7, 23.2) & 31.5 (30.6, 32.4) & \textbf{100.0 (100.0, 100.0)} \\
\midrule
\textbf{Neural A*} & \textbf{87.7 (86.6, 88.9)} &  \textbf{40.1 (38.9, 41.3)} &  \textbf{52.0 (50.7, 53.3)} & \textbf{100.0 (100.0, 100.0)} \\
\midrule
\midrule
\multicolumn{5}{c}{\textsc{Tiled MP Dataset}} \\
\midrule
  & Opt & Exp & Hmean & Suc\\ 
  \midrule
VIN & 52.7 (51.5, 54.0) & N/A & N/A & 58.3 (57.1, 59.5) \\
GPPN & \textbf{68.2 (67.0, 69.4)} & N/A & N/A & 81.5 (80.5, 82.5) \\
MMP & 44.8 (42.4, 47.1) & 40.5 (38.7, 42.4) & 35.5 (33.8, 37.2) & \textbf{100.0 (100.0, 100.0)} \\
\midrule
\textbf{Neural A*} &63.0 (60.7, 65.2) &  \textbf{55.8 (54.1, 57.5)} &  \textbf{54.2 (52.6, 55.8)}  & \textbf{100.0 (100.0, 100.0)} \\
\midrule
\midrule
\multicolumn{5}{c}{\textsc{CSM Dataset}} \\
\midrule
   & Opt & Exp & Hmean & Suc\\ 
  \midrule
VIN & 70.4 (69.3, 71.6) & N/A & N/A & 73.2 (72.1, 74.4) \\
GPPN & 68.9 (67.7 60.1) & N/A & N/A & 85.3 (84.4, 86.2) \\
MMP & 66.4 (64.0, 68.9) & 28.4 (26.4, 30.4) & 31.9 (30.0, 33.8)  & \textbf{100.0 (100.0, 100.0)} \\
\midrule
\textbf{Neural A*} & \textbf{73.5 (71.5, 75.5)} &  \textbf{37.6 (35.5, 39.7)} &  \textbf{43.6 (41.7, 45.5)} & \textbf{100.0 (100.0, 100.0)} \\
\bottomrule
\end{tabular}
}
\end{table*}

\begin{table*}[t]
\parbox{0.75\textwidth}{\caption{\textbf{Comparisons with Imitation Learning Methods on SDD.} Bootstrap means and 95\% confidence bounds of the chamfer distance between predicted and actual pedestrian trajectories.}
\label{tab:sdd_extra}}
\centering
\scalebox{0.9}{
\begin{tabular}{lcc}
    \toprule 
  & Intra-scenes & Inter-scenes\\ 
  \midrule
VIN & 920.7 (890.8, 950.4) & 900.6 (888.1, 913.8) \\
GPPN & 920.7 (890.8, 952.3) & 900.6 (888.0, 913.4) \\
MMP & 126.7 (119.3, 133.9) & 130.3 (128.1, 132.6) \\
\midrule
\textbf{Neural A*} & \textbf{16.1 (13.2, 18.8)} & \textbf{37.4 (35.8, 39.0)} \\
\bottomrule
\end{tabular}
}
\end{table*}

\subsection{Path Length Optimality Evaluation}
Following \citet{tamar2016value,anderson2018evaluation}, we introduce another metric for evaluating the path optimality, which calculates the ratio of the optimal path length $\bar{P}$ to predicted one $P$. For consistency with the other metrics, this path-length ratio is measured in 0--100 (\%) and maximized when the path is optimal, \ie, we calculate $|\bar{P}| / |P| \times 100$.
As shown in \tabref{tab:path_length}, Neural A* produced the most nearly-optimal paths across all the datasets.

\begin{table*}[t]
\parbox{.75\textwidth}{\caption{\textbf{Path Length Optimality Evaluation.} Bootstrap means and 95\% confidence bounds of the ratio of optimal to produced path lengths (the higher the better).}
\label{tab:path_length}}
\centering
\scalebox{0.80}{
\begin{tabular}{lccc}
\toprule
   & MP & Tiled-MP & CSM \\ 
  \midrule
BF & 96.4 (96.1, 96.6) & 92.1 (91.6, 92.6) & 96.4 (96.1, 96.7) \\
WA* & 96.9 (96.6, 97.1) & 93.4 (93.0, 93.8) & 96.8 (96.6, 97.1)  \\
\midrule
SAIL& 87.5 (86.8, 88.3) & 78.0 (77.0, 79.0) & 87.1 (86.1, 88.1) \\
SAIL-SL & 88.2 (87.5, 89.0) & 73.4 (72.1, 74.7) & 84.7 (83.5, 85.9) \\
BB-A* & 96.3 (96.0, 96.6) & 93.0 (92.5, 93.4) & 96.5 (96.2, 96.8)\\
\midrule
Neural BF & 97.5 (97.3, 97.8) &  95.0 (94.7, 95.4) & 97.4 (97.1, 97.6) \\
\textbf{Neural A*} & \textbf{99.1 (99.0, 99.2)} & \textbf{98.4 (98.3, 98.6)} & \textbf{98.9 (98.8, 99.0)} \\
\bottomrule
\end{tabular}
}
\end{table*}

\subsection{Computational Complexity and Runtime Analysis}
The main computational bottleneck of Neural A* lies in the differentiable A* module. Because this module uses matrix operations involving all the nodes to enable back-propagation, its training-phase complexities in space and time are $\mathcal{O}(k|\mathcal{V}|)$, where $k$ is the number of search steps. In theory, the required number of search steps depends on the true path length $d$. Thus, the complexities in terms of $d$ amount to $\mathcal{O}(d|\mathcal{V}|)$ and $\mathcal{O}(b^d |\mathcal{V}|)$ for the best and worst case, respectively, where $b$ is the average number of neighboring nodes expanded per search step. Practically, when training is finished, we can replace the differentiable A* module with a standard (\ie, non-differentiable) A* search algorithm without changing the testing behaviors of Neural A*. With this replacement, Neural A* can perform planning in $\mathcal{O}(d)$ and $\mathcal{O}(b^d)$ for the best and worst case.

To analyze search runtimes empirically, we created three sets of 50 maps with the sizes of $64\times64$, $128\times128$, and $256\times256$, by randomly drawing maps from the MP dataset and tiling them. We then measured the average runtime per problem taken using a standard A* search implementation\footnote{We implemented a python-based A* search algorithm with \texttt{pqdict} package (\url{https://github.com/nvictus/priority-queue-dictionary}) and used its priority queue feature for performing node selections efficiently. To accurately measure a search runtime per problem, we performed the program exclusively on a single CPU core (for performing A* search) and a single GPU (for additionally running the encoder to compute guidance maps for Neural A*) and solved the same problem five times after a single warm-up trial. The use of more sophisticated A* search implementations could result in further performance improvements, which is however beyond the scope of this work.} coupled with and without our guidance maps. Regardless of the test map sizes, the guidance maps were trained using the Tiled MP dataset of the size $64\times 64$, to see if our model generalizes well to larger maps. As shown in \tabref{tab:runtime}, we confirm that Neural A* greatly reduced runtimes of A* search with the help of guidance maps and also showed good generalization ability to larger maps. 

\begin{table*}[t]
\parbox{.75\textwidth}{\caption{\textbf{Search Runtime Evaluation.} Bootstrap mean and 95\% confidence bounds of the runtime (sec) required to solve a single problem with different map sizes.}
\label{tab:runtime}}
\centering
\scalebox{0.80}{
\begin{tabular}{lccc}
\toprule
   & $64\times64$ & $128\times128$ & $256\times256$ \\ 
  \midrule
  A* & 0.09 (0.08, 0.10)  & 0.21 (0.17, 0.25) & 0.78 (0.72, 0.82) \\
  Neural A* & \textbf{0.04 (0.04, 0.04)}  & \textbf{0.07 (0.06, 0.08)} & \textbf{0.15 (0.14, 0.16)} \\
\bottomrule
\end{tabular}
}
\end{table*}

\end{document}